\documentclass[runningheads]{llncs}
\usepackage{amsmath,amssymb} % define this before the line numbering.
\usepackage{color}
\usepackage{graphicx}
\usepackage[export]{adjustbox}
\usepackage{booktabs}
\usepackage{array}
\usepackage{xspace}
\usepackage{cases}
\usepackage{multirow,boldline}
\usepackage{url}
\usepackage{xcolor}
\usepackage{paralist}
\usepackage[normalem]{ulem}
\usepackage{enumitem}
\usepackage{comment}
\newcommand{\etal}{\textit{et al.}}
\newcommand{\ra}[1]{\renewcommand{\arraystretch}{#1}}
\newcommand{\mycaption}[2]{\caption{\textbf{#1.}~#2}}
\usepackage{pifont}
\newcommand{\cmark}{\ding{51}}%
\newcommand{\xmark}{\ding{55}}%
\begin{document}
\pagestyle{headings}
\mainmatter

%\def\ACCV20SubNumber{206}  % Insert your submission number here

%===========================================================
\title{Semantics-Guided Clustering with Deep Progressive Learning for Semi-Supervised Person Re-identification} % Replace with your title
\titlerunning{SG-DPL for Semi-supervised Person Re-ID}
%\authorrunning{ACCV-20 submission ID \ACCV20SubNumber}

\author{
Chih-Ting Liu\inst{1} \and
Yu-Jhe Li\inst{2} \and
Shao-Yi Chien\inst{1} \and
Yu-Chiang Frank Wang\inst{1} }

\institute{National Taiwan University, Taipei, Taiwan \\ \and
 Carnegie Mellon University, Pittburgh, PA, USA
\email{jackieliu@media.ee.ntu.edu.tw}, 
\email{yujheli@cs.cmu.edu},
\email{\{sychien,ycwang\}@ntu.edu.tw}}

\maketitle

\begin{abstract}
Person re-identification (re-ID) requires one to match images of the same person across camera views. As a more challenging task, semi-supervised re-ID tackles the problem that only a number of identities in training data are fully labeled, while the remaining are unlabeled. Assuming that such labeled and unlabeled training data share disjoint identity labels, we propose a novel framework of Semantics-Guided Clustering with Deep Progressive Learning (SGC-DPL) to jointly exploit the above data. By advancing the proposed Semantics-Guided Affinity Propagation (SG-AP), we are able to assign pseudo-labels to selected unlabeled data in a progressive fashion, under the semantics guidance from the labeled ones. As a result, our approach is able to augment the labeled training data in the semi-supervised setting. Our experiments on two large-scale person re-ID benchmarks demonstrate the superiority of our SGC-DPL over state-of-the-art methods across different degrees of supervision. In extension, the generalization ability of our SGC-DPL is also verified in other tasks like vehicle re-ID or image retrieval with the semi-supervised setting. 
\end{abstract}
\section{Introduction}
\label{sec:introduction}
Person re-identification (re-ID) aims at matching images of the same person captured by disjoint surveillance cameras, which is a challenge task owing to remarkable illumination, viewpoint and pose variation of the same pedestrian. Recently, along with the emergence of large-scale datasets~\cite{Market1501,DukeReID}, methods employing deep convolutional neural networks (CNN) have demonstrated great successes in re-ID~\cite{kalayeh2018human,PRW,hermans2017defense,zheng2016person,SVDnet,HACNN,alignedreid}. Relying on fully labeled data and state-of-the-art CNN models, very promising performances have been reported.
Yet, in practical scenarios, one might not be able to collect such a large amount of labeled data in a scene of interest for training purposes. Instead, one typically encounters semi-supervised setting in real-world re-ID tasks. More precisely, one can collect a number of fully labeled pedestrian data across camera views during specific time period, while the remaining training data under such views observed at other time periods remain unlabeled. 
Thus, one cannot easily apply and train existing supervised re-ID methods on semi-supervised data.
%While it would be desirable to perform person re-ID in a semi-supervised manner, one cannot easily apply and train existing re-ID models on semi-supervised data. 

To address the aforementioned problem of semi-supervised person re-ID, %many existing methods are proposed to tackle cross-domain unsupervised person re-ID~\cite{wang2018transferable,ARN,PUL,bak2018domain,SPGAN,InMatters}, focusing on transferring knowledge from an auxiliary source domain (with ground truth labels) to the unlabeled target domain of interest. Nonetheless, noted by Li~\etal~\cite{TAUDL}, the re-ID performance still possess a gap between the unsupervised and the fully-supervised settings on target domain due to the largely different data distribution between two domains. Recently, 
one can consider two possible settings. Recent works like~\cite{li2018semi,POE} assumes that each identity has at least one image in the training set. However, in practice, identity labels of labeled and unlabeled ones \textbf{do not} overlap (e.g., re-ID of different time periods). Thus, we follow the setting in~\cite{semi-K,icipxin} that only a small part of the identities (and their data) are seen and available. For the remaining training data, they are from a separate set of identities and are totally unlabeled during the training process. In other words, the identities of labeled and unlabeled training set are non-overlapped. %\textcolor{red}{Do you need to cite \cite{one-shot,stepwise} and where?} @liu: I can cite them in related works.

%some works turn to address the semi-supervised person re-ID~\cite{one-shot,stepwise,li2018semi,POE,semi-K}, which is more of a realistic scenario due to the limited labeling resources, while only a small part of data are labeled from the entire training dataset.

It is worth noting that, the above semi-supervised person re-ID setting is rarely addressed but practical and also challenging, since the number of identities is unknown in the unlabeled set.
With only a small part of labeled identities available in this semi-supervised setting, we need to exploit the unlabeled images to assign pseudo-labels for training purposes. Existing works like~\cite{semi-K,icipxin} simply apply K-means clustering on the unlabeled data, and then assign pseudo-labels to these data according to clustering results. However, they need to assume that the number of cluster K (i.e., identity) is known before training. They directly use the ground truth number of identities to obtain the best results, which might not be sufficiently practical either. Furthermore, assigning pseudo-labels to all unlabeled data as~\cite{semi-K} needs to be carefully handled, otherwise undesirable labeling errors would degrade the performance of the re-ID model.

To address semi-supervised person re-ID with labeled and unlabeled training data sharing disjoint identity labels, we propose a \textit{Semantics-Guided Clustering with Deep Progressive Learning (SGC-DPL)} framework. By jointly exploiting labeled and unlabeled training data, our SGC-DPL aims to augment original label information for learning re-ID models. With the guidance of labeled training data, we first advance the affinity propagation (AP)~\cite{AP} and propose the Semantics-Guided AP (SG-AP), which is a clustering technique without knowing the number of cluster K. Then, we identify and assign pseudo-labels for the unlabeled training data based on the clustering results in a progressive fashion. That is to say, we will gradually enlarge the number of unlabeled data be assigned pseudo-labels for alleviating the errors in the original clustering results. In addition, different from~\cite{icipxin,PUL}, our progressive learning approach does not require any pre-defined selection threshold or the total number of the assigned unlabeled data, which is also determined by the guidance of the labeled data.  
%With the guidance of labeled training data, we focus on identifying and assigning pseudo-labels for the unlabeled training data in a progressive fashion. As detailed later, we first advance affinity propagation (AP)~\cite{AP} and jointly exploiting labeled/unlabeled data in a progressive learning fashion. We do not need to know the number of identities in the unlabeled training set as the prior work did~\cite{semi-K}. In addition, different from~\cite{POE}, our approach does not pre-define the cardinality of the unlabeled data during the pseudo-label assignment stage.

%or the cardinality of \textcolor{red}{WHAT}~\cite{POE} as prior works did.

To the best of our knowledge, in the task of person re-ID, we are among the first to leverage the knowledge in labeled set to perform clustering without knowing the number of cluster in advance. Furthermore, we do not require heuristic hyperparameters selection in our AP-based learning model due to our jointly/iteratively exploiting labeled and unlabeled training data.

We now highlight the contributions of this work:
\vspace{-1mm}
\begin{enumerate}
\item We address the task of semi-supervised person re-ID with labeled/unlabeled training data sharing disjoint identity labels.
\item With the guidance of labeled data, our proposed Semantics-Guided Clustering with Deep Progressive Learning (SGC-DPL) framework can jointly exploit the labeled and unlabeled training data in a progressive fashion, while no prior knowledge of the number of identities and the amount of assigned unlabeled data are needed.  
%jointly exploiting labeled and unlabeled training data for augmenting labeled ones for learning re-ID models.
%\item With the guidance of labeled data, our SG-DPL progressively assign pseudo-labels to selected unlabeled training data, while no prior knowledge of the number of identities and the amount of assigned unlabeled data are needed.
\item Our model performs favorably against state-of-the-art semi-supervised re-ID approaches, and produces impressive results when comparing to fully-supervised methods.
\end{enumerate}
\vspace{-3mm}
\section{Related Work}
\vspace{-3mm}
\paragraph{\textbf{Supervised person re-ID.}}
With the recent success of deep learning, recent re-ID methods~\cite{LOMO,Market1501,zheng2016person,PRW,SPReID,hermans2017defense,DeepCC,PCB,bagoftricks,liu} rely on learning CNN models using a large number of labeled training data. Once the learning of complete, re-ID can be simply performed by matching features of query and gallery images. 
Generally, two types of loss functions are considered for training re-ID CNN networks: identity classification and verification losses. The former is viewed as the cross-entropy loss~\cite{PRW,zheng2016person}, which encourages the network to correctly recognize the identities of input images. On the other hand, popular verification loss like triplet loss~\cite{hermans2017defense,DeepCC} are utilized to encode input images, so that positive and negative image pairs can be distinguished properly in the learned embedding space. Recent works like~\cite{bagoftricks} jointly use these two types of losses, and very promising results are reported. As noted above, while these methods achieve promising re-ID performance, they require a large amount of labeled data for training purposes, which is often not practical in real-world re-ID applications.

\paragraph{\textbf{Semi-supervised person re-ID.}}
Since collecting and annotating a large amount of training data are often not applicable in real-world applications, how to design and train re-ID models in a semi-supervised setting would be of increasing interest. 
Some works~\cite{GAN-1,GAN-2} approach this setting by utilizing labeled training data for synthesizing unlabeled ones via Generative Adversarial Network (GAN)~\cite{GAN}. Once the synthesized images are generated, the multi-pseudo regularized labels can be assigned like~\cite{GAN-1} or the labels are determined according to the relation of labeled and unlabeled data in the feature space~\cite{GAN-2}. However, the generated data are not visually robust and the real unlabeled data are also neglected for training the netowrk.
%Huang~\etal~\cite{GAN-1} assign a multi-pseudo regularized label, and Ding~\etal~\cite{GAN-2} assign the pseudo labels by considering the relation of labeled and unlabeled data in the feature space. With synthesized data and the associated labels, the supervised re-ID models can be trained accordingly.
A number of works focus on one-example or few-example settings~\cite{li2018semi,stepwise,one-shot,POE}, i.e., assuming that only one or few images of each identity are available in the training set, while the remaining ones are of the same identities but unlabeled during training. In~\cite{li2018semi}, a region metric learning method is proposed, which identifies neighbors of the same identity labels and forms a discriminative metric. Wu~\etal~\cite{POE} propose a learning method for the unlabeled data which contains the exclusive loss and a progressive pseudo-labels estimation technique. While the above setting requires semi-supervised learning models, one might not be able to collect labeled data for each identity in advance and cannot expect that the identities in unlabeled data would remain the same.
%that progressively select a pre-defined number of data from the unlabeled set based on the one-example labeled training set. While the above setting requires semi-supervised learning models, one might not be able to collect labeled data for each identity in advance. When deploying and learning re-ID models across scenes or data domains, one cannot expect that the identity labels would remain the same.

As an alternative and possibly more practical semi-supervised setting,~\cite{PUL,semi-K,icipxin} considers that only a small number of identity labels and their image data are observed in advance, while the remaining data are unlabeled and with \textit{distinct} labels (i.e., ground truth labels of labeled and unlabeled datasets do no overlap). A standard k-means algorithm is applied in~\cite{PUL,semi-K,icipxin} for assigning pseudo-labels for unlabeled data, requiring the prior knowledge of the number of clusters K (i.e., the number of identities). Although they demonstrate that the performance is robust in a range of K, we also need to determine the suitable range. In this work, we propose to jointly observe labeled and unlabeled training data during learning. While ground truth labels and the number of identities in unlabeled training data are never observed, our \textit{semantics-guided clustering with deep progressive learning} allows us to identify and assign pseudo-labels to selected unlabeled data progressively, so that satisfactory re-ID performance can be preserved.

\paragraph{\textbf{Semi-supervised affinity propagation and Progressive learning}}
Without knowing the number of clusters in advance, affinity propagation (AP) is a suitable clustering solution. Some works~\cite{semi-ap-1,semi-ap-2} propose semi-supervised AP that utilizes the labeled data as additional constraints when clustering on the unlabeled data. However, both constraints assume a shared label space between those labeled and unlabeled data, which is not suitable for real-world settings. Our proposed semantics-guided AP can learn an adaptive AP mechanism on the labeled set and adapt it to the disjoint unlabeled set. 

Progressive learning, which is in the field of self-paced learning (SPL)~\cite{spl}, aims to obtain knowledge from easy to hard samples in a pre-defined scheme, and the self-paced paradigm is theoretically analyzed in~\cite{spl-1,spl-2}. In the semi-supervised re-ID field, many works~\cite{PUL,POE,icipxin} adopt the progressive learning scheme but they all need to determine a heuristic parameter for the selection threshold or the cardinality of the unlabeled set for training between each iteration. In this paper, we exploit the labeled data and propose a progressive learning method that can automatically generate the suitable threshold for data selection. 

\paragraph{\textbf{Unsupervised person re-ID.}}
We note that, a number of unsupervised person re-ID works are presented. Approaches like~\cite{Market1501,LOMO} devise hand-crafted features to represent pedestrian images for matching purposes. %leading to inferior performance due to the lack of ability to handle cross-camera image variations. 
Some works like BUC~\cite{buc} and AE~\cite{exploration} try to directly learn discriminative CNN representations on the target unlabeled dataset. For example, Lin~\etal~\cite{buc} utilize bottom up clustering to leverage the pseudo-labels, while Ding~\etal~\cite{exploration} adaptively select image pairs for training re-ID. On the other hand, most works choose to learn the representation of the unlabeled data with the aid of a source dataset in the other domain. MAR~\cite{SML} propose the soft-multilabel technique for the data in target domain which is based on the relation of the unlabeled data to all the labeled source identities, while SSG~\cite{SSG} utilize a self-similarity grouping to mine the potential similarities for both global and local features. Recently, MMT~\cite{mutual} proposed a framework to off-line refine hard pseudo-labels and on-line refine soft pseudo-labels in an alternative training manner. Since annotating at least a small amount of data in the target domain is practical for real-world re-ID applications, we will focus on the semi-supervised setting as described above and will not address the pure or cross-domain unsupervised settings.
\section{Semantics-Guided Clustering with Deep Progressive Learning for Semi-Supervised Person Re-ID}
\vspace{-1mm}
For the sake of completeness, we first define the problem formulation of semi-supervised re-ID and the notations used in this paper.
Assume that we have access to a set of $N^l$ labeled images $X^l= \{ x_{i}^{l}\}_{i=1}^{N^l}$ and their associated label set $Y^l=\{y_{i}^{l}\}_{i=1}^{N^l}$, where $y_{i}^{l} \in [1,2,...,C^l]$ and $C^l$ denotes the number of identities in the labeled data. In addition, another set of $N^u$ images $X^u= \{ x_{j}^{u}\}_{j=1}^{N^u}$ without any label information are also available during training. Note that the number of identities $C^u$ in the unlabeled set $X^u$ is \textit{unknown} (which is different from~\cite{semi-K,icipxin}), while their identities are \textit{non-overlapped} with $Y^l$. 

\begin{figure*}[t]
	\centering
    \includegraphics[width=\textwidth]{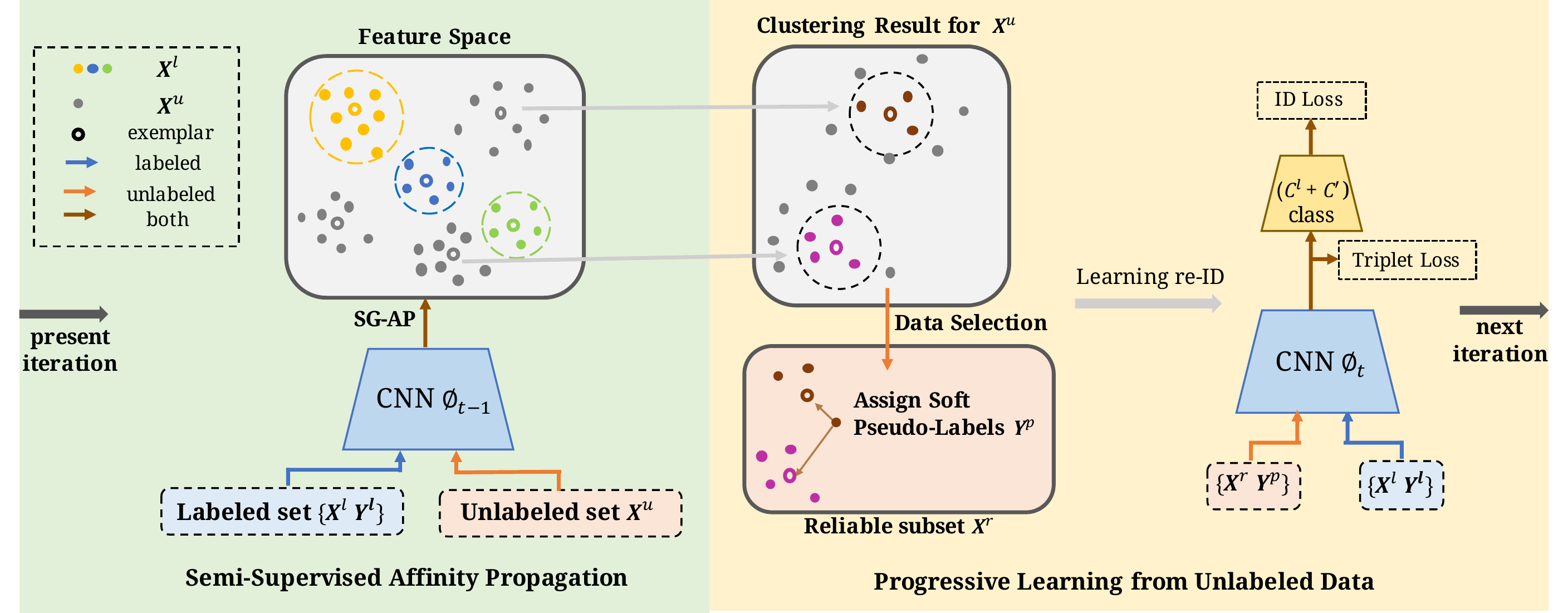}
    \mycaption{Overview of our proposed SGC-DPL for semi-supervised re-ID}{At each iteration $t$, we perform semantics-guided affinity propagation (SG-AP) to jointly cluster labeled and unlabeled data and progressively select a subset from unlabeled data for soft pseudo-label assignment. This augments labeled dataset without knowing the exact number of ID labels in advance.}
    \label{fig:big}
\end{figure*}

Instead of training the CNN model using \{$X^l$,$Y^l$\} only, we additionally leverage the image from $X^u$ to augment the labeled training data. As depicted in Fig.~\ref{fig:big}, we propose \textit{Semantics-Guided Clustering with Deep Progressive Learning (SGC-DPL)} for solving this semi-supervised person re-ID task. This is realized by our semantics-guided affinity propagation (SG-AP) and progressive data selection strategies. This would iteratively assign soft pseudo-labels $Y^p$ to a selected subset $X^r \subset X^u$, and augment labeled data for training standard re-ID models.
\vspace{-3mm}

\begin{figure*}[t]
	\centering
    \includegraphics[width=0.7\textwidth]{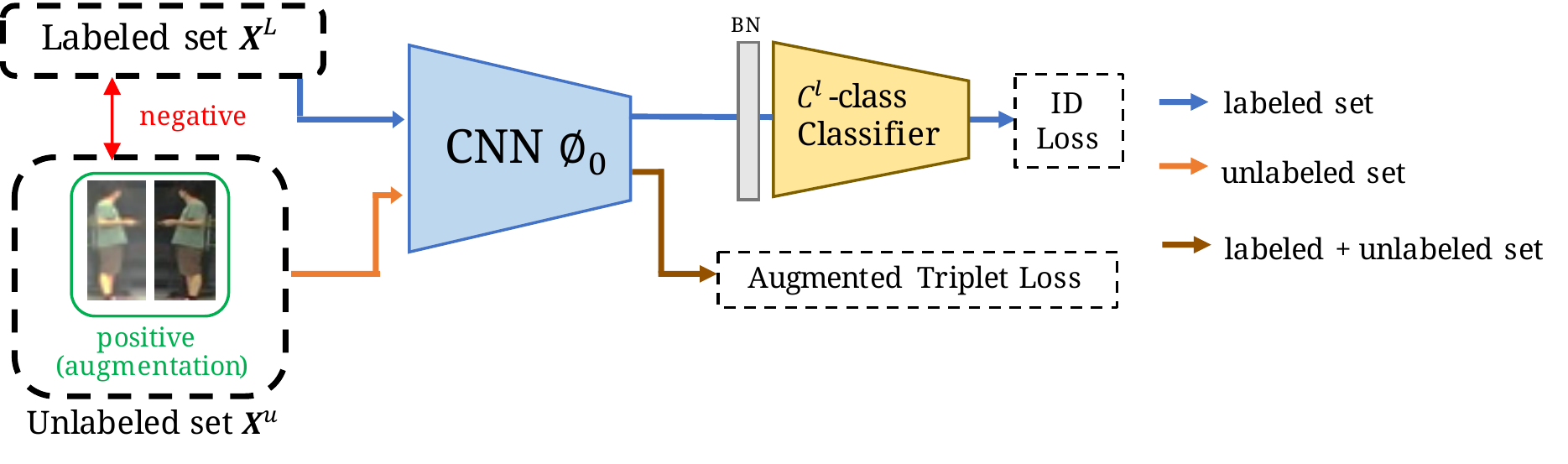}
    \mycaption{Model initialization for semi-supervised re-ID}{To initialize the re-ID model, the ID/triplet losses are observed from $\{X^l,Y^l\}$, while the augmented triplet loss is additionally observed by exploiting positive pairs from $X^u$ and negative pairs across $X^l$ and $X^u$.}
    \label{fig:initial}
    \vspace{-4mm}
\end{figure*}

\subsection{Model Initialization in Semi-Supervised Re-ID}
\label{sub:initial}
We now present our model initialization process, which is depicted in Fig.~\ref{fig:initial}. Following the model architecture and the training strategy described in~\cite{bagoftricks}, we first use a CNN as a feature extractor $\phi$, and thus the features of labeled images $\phi(x^l)$ are used to train the batch-hard triplet loss~\cite{hermans2017defense}. A BatchNorm~\cite{bn} and a fully-connected layer are used to construct a $C^l$-class classifier for optimizing the identity classification loss (ID loss)~\cite{PRW}. 

In our semi-supervised re-ID task, ID labels are non-overlapped between training data $X^l$ and $X^u$. Inspired by~\cite{HHL}, we further propose an \textit{augmented triplet loss} that utilize the unlabeled set to generate additional positive and negative pairs. To be more specific, given any image in $X^u$, we first perform data augmentation for an unlabeled image as a novel image with the same label (and thus form a positive pair). On the other hand, we also randomly pick any two images from $X^l$ and $X^u$ (one from each) to form a negative pairs. Therefore, the original triplet loss will be observed by such augmented positive and negative pair data. 

\vspace{-3mm}
\subsection{Semi-Supervised Affinity Propagation}
\label{sub:AP}
Without knowing the number of clusters in advance, Affinity Propagation (AP)~\cite{AP} is a robust unsupervised clustering algorithm, which is analyzed in our supplementary materials with DBSCAN~\cite{DBSCAN}. To jointly exploit labeled and unlabeled training data for learning re-ID models, we present a novel algorithm of semantics-guided affinity propagation (SG-AP), which is a semi-supervised clustering method. Based on AP, we additionally perform clustering on labeled data to generate semantics (i.e., ID label) guidance for clustering on unlabeled set. That is, we aim at preserving the consistency between the clustering and identity outputs, and augment labeled data from the unlabeled data set for semi-supervised training purpose. Next in Sec.~\ref{subsub:pre-AP}, we will briefly review AP algorithm followed by our proposed semantics-guided affinity propagation described in Sec.~\ref{subsub:semanticAP}.

\vspace{-2mm}
\subsubsection{Brief Review of Affinity Propagation}
\label{subsub:pre-AP}
Given a set of unlabeled data points $X=\{x_1,~x_2, ..., x_N\}$, AP takes one similarity matrix $s$ between data points as input, where each similarity element $s(i,j)$ shows how likely $x_j$ would serve as an exemplar for $x_i$. The similarity score can be calculated via $s(i,j)=-{\lVert \phi(x_i)-\phi(x_j) \rVert }^2_2$, where $i \neq j$. This formula indicates the negative euclidean distance between feature points. Note that this distance metric is concurrently optimized by triplet loss in re-ID task, which is also beneficial to the clustering result. %and is applicable to triplet losses in re-ID tasks.
\textit{Without} pre-defining the number of objective clusters, AP only needs to define a score $s(i,i)$ for each data point $i$ so that data points with larger $s(i,i)$ are more likely to be chosen as cluster exemplars. These values are called ``\textbf{\textit{preferences}}". Such preferences will greatly affect the final clustering result after the learning procedure of AP. However, it is hard to decide the proper preference value for each data point while the value is usually given based on heuristic experiments. In the original AP~\cite{AP} algorithm, the preference values are ``equally'' assigned to all the data as $s(i,i)=p~~ \forall i$, where $p$ is either set to be as the median of the pairwise similarities, which results in a moderate number of clusters, or their minimum resulting in a small number of clusters. %If the larger the preference is, the more the final number of clusters will be.
To be more precise of AP learning procedure, two values are passed between data points during internal clustering iteration: responsibility $r$ and availability $a$. For each step $t$, responsibility $r_t(i,j)$ is calculated by the similarity matrix $s$ and $a_{t-1}$, and availability $a_t(i,j)$ is calculated by $r_{t-1}$. Finally, for a data point $x_i$, the exemplar of $x_i$ is selected by:
%For each step $t$, responsibility $r_t(i,j)$ is calculated by the similarity matrix $s$ and $a_{t-1}$, which shows how likely a point $j$ is to serve as the exemplar for another point $i$, while taking into account other potential exemplars for $i$. On the other hand, availability $a_t(i,j)$ is calculated by $r_{t-1}$, which reflects the accumulated confidence showing how likely the point $i$ chooses $j$ as its exemplar, taking into account the support from other points that point $j$ should be an exemplar. Finally, for a data point $x_i$, the exemplar of $x_i$ is selected by:
\begin{equation}
c_i \leftarrow \mathop{\arg\max_{x_j}}\{r(i,j) + a(i,j)\},
\end{equation}
where $c_i$ denotes the exemplar for $x_i$ when convergence. 

\vspace{-2mm}
\subsubsection{Semantics-Guided Affinity Propagation}
\label{subsub:semanticAP}
While AP is an effective unsupervised clustering algorithm not requiring the prior knowledge of the number of clusters, it cannot be directly applied to semi-supervised re-ID tasks. This is because that performing clustering on the unlabeled dataset does not necessarily output data clusters corresponding to desirable ID labels. Moreover, assuming all data points possess the same preference with value $p$ hinders the clustering results. To overcome the above challenges, we present \textit{semantics-guided affinity propagation (SG-AP)}, which jointly exploit labeled and unlabeled training data. With the semantics (i.e. ID label) guidance of labeled data, our goal is to cluster and assign psuedo-labels for unlabeled ones to augment the labeled data for training purposes.

To solve the aforementioned problem that preference values of all data are equally assigned, %To better cluster unlabeled data, 
our SG-AP first introduces an adaptive preference function that generates a suitable preference of each data point based on the observed feature distribution, which is produced by calculating the similarities between each data point to the others. The core idea is that, if the distance between a point $x_i$ to other points is larger than the one between $x_j$ to others, the point $x_i$ should has a lower possibility to be a cluster exemplar than $x_j$ does, which results in a lower preference value. To achieve this goal, we first define the Similarity Ranking coefficient ($SR$) of each $x_i$ as:
\begin{equation}
    SR(x_i) = N \times \frac{\sum_{j=1,j\neq i}^{N}{s(i,j)}}{\sum_{i=1}^{N}{\sum_{j=1,j\neq i}^{N}{s(i,j)}}},
\end{equation}
where $N$ is the number of clustered data and $s(i,j)$ indicates the element in the similarity matrix $s$ of the data points. The summation of the similarities among data point $x_i$ to other $N$ points ($\sum_{j=1,j\neq i}^{N}{s(i,j)}$) will be normalized and multiplied by $N$ to represent the relative ranking value of $x_i$ be chosen as a cluster exemplar among the $N$ data points. Then, we can define the adaptive preference of $x_i$ as:
\begin{equation}
\label{eq:sim_AP}
   Adaptive~Preference(x_i) = s(i,i) = SR(x_i) \times p,
\end{equation}
where the $SR(x_i)$ serves as an adaptive ranking weight for the original preference $p$, resulting in different preference values for each data point based on its similarities to the other data points. Note that both the elements in similarity matrix $s$ and $p$ are negative values; therefore, a data point with high relative ranking to be a cluster exemplar will result in a smaller $SR$ and a larger preference $s(i,i)$.

Although we have adaptive preference, the constant $p$ is still determined by heuristic experiments (median or minimum of similarities), and that might lead to undesirable cluster results on the unlabeled data. To exploit the semantics information (ID labels) in the labeled set, we proposed our SG-AP in the semi-supervised manner. 
%It is worth noting that, in our SG-AP, $p$ is determined by observing labeled data in the semi-supervised setting. 
This is realized by enforcing the clustering of labeled data to fit the desirable ID labels. That is, given labeled and unlabeled data $\{X^l,X^u\}$, we first calculate~\eqref{eq:sim_AP} with $p$ initially set as the median of similarity matrix observed from the labeled set, where $N$ equals to $N^l+N^u$. Since the number of identities $C^l$ is known, we can search for $p^*$ that makes the number of exemplars of labeled set after clustering best matches $C^l$. If the number of exemplars is larger than $C^l$ (i.e., over-clustering), smaller $p$ will be considered (and vice versa). This searching process can be sped up with Binary Search on data pair similarities observed from $X^l$. With $p=p^*$ and $N=N^u$ in~\eqref{eq:sim_AP}, we perform clustering on $X^u$ and obtain $C^{\prime}$ exemplars, and such results are guided by the semantics information observed in $X^l$ as described above.

\vspace{-3mm}
\subsection{Progressive Learning from Unlabeled Data}
\label{sub:prog}
Our SG-AP performs clustering on unlabeled data based on the semantics guidance of labeled dataset. To jointly exploit labeled and unlabeled data for training effective re-ID models, the second stage in our SGC-DPL is to progressively assign soft pseudo-labels for selected unlabeled data with high confidence, so that learning of semi-supervised re-ID models can be further achieved. 
\vspace{-3mm}
\subsubsection{Progressive Data Selection Strategy}
\label{subsub:data_selection}
To better leverage the clustering results after our SG-AP process, we now present a data selection strategy by choosing a reliable subset $X^r$ from the unlabeled set $X^u$ in a progressive fashion, as shown at the right part of Fig.~\ref{fig:big}. For each cluster, if the instances $x^u_i$ of that cluster whose feature-level distance to the exemplar $x^u_{c_i}$ is smaller than a threshold $\tau$, we will select such instances with the corresponding labels into the reliable subset $X^r$. The threshold $\tau$ will be progressively enlarged to bring in more unlabeled data to effectively train the re-ID model. A formal definition for $X^r$ can be formulated as follows:
\begin{equation}
\label{eq:reliable}
   X^r = \{x^u_i \mid ~{\lVert \phi(x^u_i)-\phi(x^u_{c_i}) \rVert }^2_2 < \tau \}
\end{equation}

\begin{figure}[t]
	\centering
    \includegraphics[width=0.4\textwidth]{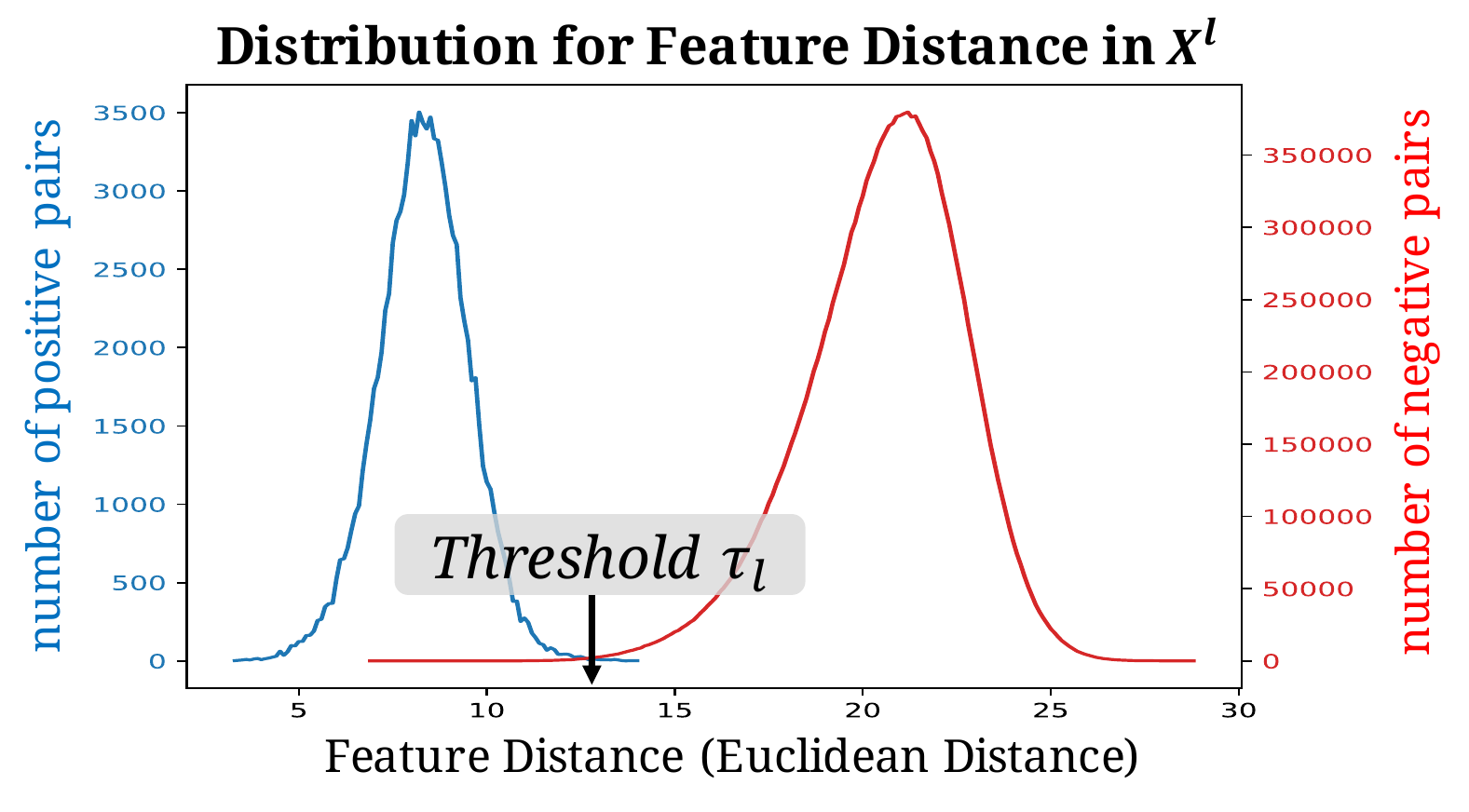}
    \mycaption{Determining threshold $\tau_l$ for progressive data selection}{We illustrate the distributions of distance between pairwise data of $X^l$ on Market-1501 with semi-supervised setting. The blue and red curves are those for positive and negative pairs, respectively. The intersection of the two curves indicates the threshold $\tau_l$ which minimizes the data assignment errors for that dataset.}
    \vspace{-3mm}
    \label{fig:distance}
\end{figure}

It is worth noting that, different from most existing progressive learning strategies which typically utilize pre-defined thresholds for data selection~\cite{PUL,POE,icipxin}, our threshold $\tau$ can be observed from the labeled set directly. To be more specific, $\tau=\tau_l + d_t$, where $\tau_l$ is determined based on labeled set, which dominates the threshold value and $d_t$ is for enlarging the threshold gradually based on the SGC-DPL iteration. Since $\tau_l$ is seen as the expected maximum distance between an exemplar and its positive members, we utilize the distance distribution of data pairs in the labeled set to leverage informative hints within such data selection process. Fig.~\ref{fig:distance} depicts the distributions of feature distances within positive and negative pairs in the labeled set $X^l$ on semi-supervised Market-1501 dataset~\cite{Market1501}. From Fig.~\ref{fig:distance}, it is obvious that we can pick a threshold which is data-dependent and separates positive and negative pairs with minimum errors. With this observation, the threshold $\tau_l$ can be assigned as the distance value on the intersection line, using the labeled training data of interest. Then, between each progressive learning iteration in SGC-DPL, $\tau_l$ will be gradually increased till all the instances are selected into the reliable set accordingly.
\vspace{-3mm}

\subsubsection{Soft Pseudo-label Assignment}
\label{subusb:soft}
To train our re-ID model in this semi-supervised setting, the above process allows us to select reliable data $X^r$ based on SG-AP results. In order to assign pseudo-labels $Y^p$ for such data without the prior knowledge of cluster/ID numbers, we choose to assign soft pseudo-labels to alleviate possible clustering or label assignment errors. That is, given a data point $x_i^r$ in $X^r$ and $C^{\prime}$ cluster exemplars, the soft pseudo-label vector $y^p_i$ is defined as follows:
\begin{equation}
\label{eq:pseu}
    y^p_i = softmax([-d(i,1),-d(i,2),...,-d(i,C^{\prime})])
\end{equation}
where $d(i,j)$ is the feature distance for data $x^r_i$ to the $j^{th}$ exemplar in the unlabeled set. In other words, for $x^r_i$, the logit of the $j^{th}$ element in $y^{p}_{i}$ depends on the distance between $x^r_i$ and the $j^{th}$ exemplar. The smaller the distance is, the larger the logit is. 

After obtaining the reliable data and its soft pseudo-labels $\{X^r,Y^p\}$, such augmented data will be added to the original labeled set $X^l$ for jointly learning for re-ID model. With refined model, the resulting feature extractor will be utilized for SG-AP and progressive data selection in the next iteration.

\vspace{-3mm}
\subsection{Learning Objective of Our Model}
\label{sub:alg}
To train our entire SGC-DPL framework for achieving semi-supervised re-ID, we alternate between the above SG-AP and progressive data selection process for assigning soft pseudo-labels to unlabeled data, which augment the original training set $\{X^l,Y^l\}$ to an updated one $\{X^l,Y^l,X^r,Y^p\}$. We then re-fine our model with the new training data in that iteration by jointly optimizing batch-hard triplet loss and ID loss as~\cite{bagoftricks}. Since new $C^{\prime}$ identities are added to the original training set, the classifier in our re-ID model will be expanded to train the ID loss with $Y^p$ and $Y^l$, where $y_i^p$ is a soft label vector used in the cross-entropy loss. In addition, our model is trained using the triplet loss. Since data pairs in $\{X^r\}$ are unlabeled, we determine the identity for selecting positive and negative pairs of $x^r_i$ by our SG-AP clustering results.
\vspace{-3mm}
\section{Experiments}
\vspace{-3mm}
\subsection{Datasets}
We evaluate our method on two benchmarks, Market-1501~\cite{Market1501} and DukeMTMC-reID~\cite{DukeReID}, which are two large-scale datasets with multiple cameras.
\\
\textbf{Market-1501}. The Market-1501~\cite{Market1501} is composed of 32,668 labeled images of 1,501 identities collected from 6 camera views. The dataset is split into two fixed parts: 12,936 images from 751 identities for training and 19,732 images from 750 identities for testing. During testing phase, 3368 query images from 750 identities are used to retrieve people in the gallery set.
\\
\textbf{DukeMTMC-reID.}  The DukeMTMC-reID~\cite{DukeReID} is a subset of DukeMTMC~\cite{Duke}, which is created for re-ID purpose. It is collected from 8 cameras and contains 36,411 labeled images belonging to 1,404 identities. 702 identities with 16,522 images are used for training, and 2,228 images from other 702 identities are used for query images retrieving the rest 17,661 gallery images.
\vspace{-3mm}

\subsection{Experimental Settings and Protocols}
We employ the standard metrics of the cumulative matching curve (CMC) and the mean Average Precision (mAP). We report the rank-1 accuracy in CMC and the mAP for the testing set in both datasets. We follow the semi-supervised settings in~\cite{semi-K,icipxin}, which splits the training set into two parts: one is labeled and the remaining is unlabeled, according to the proportion ratio of person identities.  The ratios are set as 1/3, 1/6, and 1/12. For example, for the 1/6 case, only about 125 among 751 identities in the training set of Market-1501~\cite{Market1501} are labeled across cameras and the remaining images in the training set are unlabeled. For fair comparison to some state-of-the-arts, we also adopt the setting that only 50 identities (50 ID) are labeled. % That is, the unlabeled dataset which contains 691 identities is unknown of its identity information.
\vspace{-3mm}
\subsection{Implementation Details}
We employ ResNet-50~\cite{resnet} as the backbone in our feature extractor $\phi$. The 2048-d feature vectors produced by last layer of our feature extractor are used for re-ID and trained with batch-hard triplet loss as well as the $PK$ training strategy suggested by Hermans~\etal~\cite{hermans2017defense}. We sample $P=16$ different identities and $K=4$ images for each person at a time to form a batch data of size 64. To improve the supervised training performance, we also follow some of the tricks proposed in~\cite{bagoftricks}, which contains the BNNeck, warmup and the REA. We use Stochastic Gradient Descent (SGD) to optimize our model $\phi_t$ for total 200 epochs with the augmented training set $\{X^l,Y^l,X^r,Y^p\}$ and with the initial learning rate of 0.01 decaying by 10 every 50 epochs. 
%As noted by Luo~\etal~\cite{bagoftricks}, we add a BatchNorm layer~\cite{bn} and a fully-connected layer after the feature extractor to optimize the identity classification loss~\cite{PRW}. 
The total training iterations of our SGC-DPL framework is set as $t=8$ . In the internal semantics-guided affinity propagation (SG-AP), the searching process of $p^*$ will be terminated if the clustered results on $X^l$ match the number of identities ($C^l$) or converge to a fixed number of exemplars for 5 iterations. In our progressive data selection, the $d_t$ is initially set as $0$ and gradually added with a step size $1$ to bring in  more unlabeled data.%For the progressive data selection strategies, the threshold $\tau_l$ will be increased by enlarged to $\tau_l+m$ depends on the number of iterations, which is set as $m=0$ for first $2$ iterations, $m=1$ from $3^{th}$ to $5^{th}$ iteration and $m=\infty$ for the last few iterations, respectively.

%LOMO~\cite{LOMO}                    & \multirow{ 2}{*}{\specialcell{pure \\unsupervised}}             & 27.2             & 8.0           \\
%BOW~\cite{Market1501}               &               & 35.8             & 14.8          \\
\newcommand{\specialcell}[2][c]{%
  \begin{tabular}[#1]{@{}c@{}}#2\end{tabular}}
\begin{table}[t]
    \centering
    \caption{\textbf{Comparisons with unsupervised and semi-supervised re-ID methods on Market-1501 and DukeMTMC-reID(\%).}}
    \label{tab:sota}
    \ra{1}
    \begin{tabular}{l|c|cc|cc}
    \hline
    \multirow{ 2}{*}{Method}   & \multirow{ 2}{*}{Supervision}  & \multicolumn{2}{c|}{~~Market-1501~~} &\multicolumn{2}{c}{DukeMTMC-reID} \\
    \cline{3-6}
    & &  Rank-1   & mAP  &  Rank-1   & mAP  \\ \hline \hline
    BUC~\cite{buc} &   \multirow{ 2}{*}{\specialcell{purely \\unsupervised}}  & 66.2  & 38.3  & 47.4  & 27.5 \\
    AE~\cite{exploration} &                                                 & 77.5  & 54.0  & 63.2  & 39.0  \\
    \hline
    MAR~\cite{SML} & \multirow{ 3}{*}{\specialcell{cross-domain \\unsupervised}}  & 67.7 & 40.0 & 67.1& 48.0 \\
    SSG~\cite{SSG}               &                                                    & 80.0 & 58.3 &73.0 & 53.4 \\
    MMT~\cite{mutual}               &                                                    & 87.7 & 71.2 & 78.0 & 65.1 \\
    \hline
    POE~\cite{POE} & one-example                           & 55.8 & 26.2 & 48.8 & 28.5  \\
    \hline \hline
    \multicolumn{6}{c}{\textbf{ID-disjoint semi-supervised}} \\
    \hline
    UMDL~\cite{UMDL}      &  \multirow{ 5}{*}{\specialcell{50 ID \\ labeled}}  & 35.6 & 13.4 & 19.5 &8.3 \\
    PUL~\cite{icipxin}      &                                  & 50.9 & 24.8 &36.5 & 21.5\\
    MVC~\cite{semi-K}      &                                    &49.9 & 24.9 &35.7 &22.5 \\
    MVSPC~\cite{icipxin}      &                                  & 62.1 & 40.9 &51.5 &31.5  \\
    \textbf{Ours}  &                            & \bf{83.8} & \bf{65.3} & \bf{74.4} & \bf{56.1}  \\
    \hline
    MVC~\cite{semi-K} & \multirow{ 3}{*}{\specialcell{1/3 ID \\ labeled}} & 75.2  & 52.6 &57.6 &37.8  \\
    MVSPC~\cite{icipxin}      &                                          & 80.1  & 62.8 &70.8  &50.3   \\
    \textbf{Ours}       &                     & \textbf{91.1}    & \textbf{76.4} & \textbf{82.2} & \textbf{66.5}\\ 
    \hline \hline
    BoT~\cite{bagoftricks} & fully-supervised                   & 94.5    & 85.9 & 86.4 &76.4\\
    \hline
    \end{tabular}
    \vspace{-3mm}
\end{table}

\vspace{-3mm}
\subsection{Evaluation}
We first compare our methods with existing two unsupervised settings, one-example setting and the fully-supervised approaches, and report the results on the two datasets in Table~\ref{tab:sota}. For the purely unsupervised methods~\cite{buc,exploration}, which directly exploit the target unlabeled data without utilizing a source dataset, there is still a performance gap to the fully-supervised method~\cite{bagoftricks} because they cannot learn the cross-camera image variation in the dataset. AE~\cite{exploration} achieve a great improvement because they additionally utilize a style transfer network provided by~\cite{camstyle}, which implies that generating various positive pairs in the domain of interest would help. For the cross-domain unsupervised methods~\cite{SML,SSG,mutual}, they can initialize and update the model with the aid of all labeled data in source dataset; therefore, the performance can be more satisfactory when comparing to the methods above.  We note that, while the one-example setting POE~\cite{POE} adopt a common semi-supervised setting and they also utilize the progressive learning to leverage reliable unlabeled data, a significant performance gap between theirs and fully-supervised BoT~\cite{bagoftricks} is also observed. This indicates that this one-example semi-supervised learning approach cannot produce promising performance in such settings.

%We compare our proposed method with existing unsupervised, semi-supervised, and fully-supervised approaches, and report the results on \textbf{Market-1501} and \textbf{DukeMTMC-reID} in Tables~\ref{tab:sota_market} and~\ref{tab:sota_duke}, respectively. From the results shown in these tables, we see that hand-crafted features used in LOMO~\cite{LOMO} and BOW~\cite{Market1501} cannot be applied to unlabeled data well.
%which typically can be adopted to any new unlabeled environment, explicit large performance drop by a large margin. This is due to their lack of training procedure as learning based models.
%Recent unsupervised deep models like BUC~\cite{buc}, DBC~\cite{dispersion} and AE~\cite{exploration} adopt CNN for feature extraction but not considering cross-camera image variations. Thus, their performance drop can be expected due to errors in predicting pseudo-labels for unlabeled data in such unsupervised settings. AE~\cite{exploration} utilize a style transfer network provided by~\cite{camstyle}, which implies that generating positive pairs in the domain of interest would help. We note that, while the one-example setting POE~\cite{POE} is the most recent work adopting progressive learning to leverage reliable unlabeled data, a significant performance gap between theirs and fully-supervised PCB~\cite{PCB} is observed. This indicates that this one-example semi-supervised learning approach cannot produce satisfactory performance in such settings.

On the other hand, as the setting considered in~\cite{UMDL,PUL,semi-K,icipxin}, our semi-supervised person re-ID utilizes \textbf{disjoint} identities in labeled and unlabeled set. For fair and complete comparisons, we only report results with 50 labeled identities and the ratio of labeled identities as 1/3 in Table~\ref{tab:sota}. Other different ratios of labeled data (1/6 and 1/12) are reported in Table~\ref{tab:ablation}. For the setting that only 50 identities are labeled, it is clear that our SGC-DPL performed against the state-of-the-art MVSPC~\cite{icipxin} by a large margin such as $\mathbf{24.4}\%$ and $\mathbf{24.6}\%$ in terms of mAP on Market-1501 and DukeMTMC-reID, respectively. The promising results can also be observed in the setting that 1/3 of the identities are labeled. Our superiority over these state-of-the-art approaches demonstrates the effectiveness of the proposed SG-AP  and the guided progressive learning against the K-means clustering without guidance and the pre-defined self-paced learning in~\cite{icipxin}. We then further analyze the effectiveness of each component in the next section.

%On the other hand, as the setting considered in MVC~\cite{semi-K}, our employed semi-supervised person re-ID setting utilizes disjoint identities in labeled and unlabeled set. MVC~\cite{semi-K} considered the ratio of labeled identities as 1/3, 1/6, and 1/12, but only reported their results of 1/3 in both rank-1 and mAP (other two are reported only in rank-1). Yet, we report results across all ratios (1/3, 1/6, and 1/12) by our proposed SG-DPL framework in the above tables. It is clear that we performed against MVC (1/3)~\cite{semi-K} by large margins such as $15.9\%$, $24.6\%$ in terms of rank-1 and $23.8\%$, $28.7\%$ in terms of mAP on Market-1501 and DukeMTMC-reID, respectively. Our superiority over these state-of-the-art approaches demonstrates the effectiveness of the proposed algorithm in semi-supervised re-ID. We then further analyze the effectiveness of each component in the next section.
\vspace{-3mm}
\subsection{Ablation Studies and Visualization}
\label{sub:ablation}
\begin{figure}[t]
	\centering
    \includegraphics[width=\textwidth]{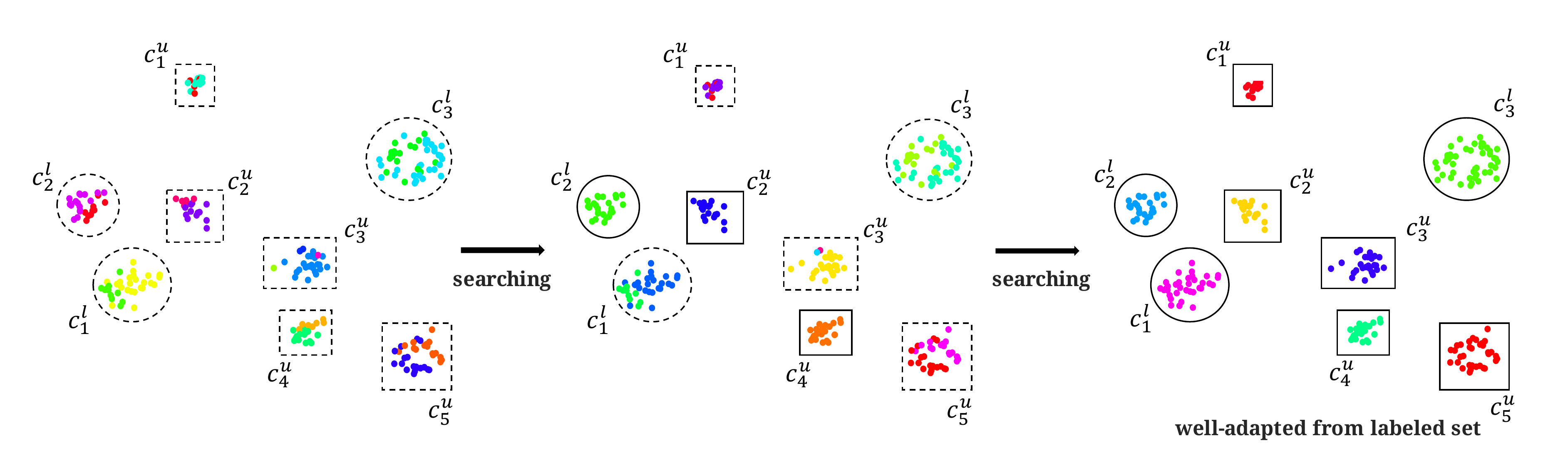}
    \mycaption{2D t-SNE visualization of internal SG-AP clustering results on sampled $X^l$ and $X^u$ from the M-1/6 dataset}{Data with the same color represent instances of the same \textit{cluster}, while labeled/unlabeled data with the same ground truth identity are bounded by circles/rectangles. Note that instances bounded by dotted circles/rectangles indicate mismatch between clustering and ID labels, while those by solid circles/rectangles denote the match between them.}
    \label{fig:SG-AP}
    \vspace{-4mm}
\end{figure}

To assess the effectiveness of each introduced component in our SGC-DPL, we conduct ablation studies and report the results on Market-1501 in Table~\ref{tab:ablation}. The experiment is composed of three kinds of semi-supervised settings, which depends on the ratio of labeled identities (1/3, 1/6 or 1/12) on Market-1501 dataset (M) and thus denoted as M-1/3, M-1/6, and M-1/12. The results on DukeMTMC-reID dataset are also reported in our supplementary materials. In addition, to demonstrate the effectiveness of our SG-AP, we visualize the clustering results along the searching process in Fig~\ref{fig:SG-AP}.

\begin{table}[t]
    \centering
    \mycaption{Ablation studies of the proposed method in terms of R-1 and mAP (\%)}{Note that Init., Clus., P.L. and Pseu.-labels indicate the uses of techniques discussed in Sec.~\ref{sub:initial}, Sec.~\ref{sub:AP}, Sec.~\ref{subsub:data_selection} and Sec.~\ref{subusb:soft}. All methods in this table share the same backbone model.}
    \label{tab:ablation}
    \ra{1}
    \begin{tabular}{l|cccc|cc|cc|cc}
    \hline
    \multirow{2}{*}{Experimental setting} & \multicolumn{4}{c|}{Components} & \multicolumn{2}{c|}{M-1/12}&\multicolumn{2}{c|}{M-1/6}&\multicolumn{2}{c}{M-1/3} \\
    \cline{2-11}
     & Init. & Clus. & P.L. & Pseu.-labels & R-1 & mAP & R-1 & mAP & R-1 & mAP \\
    \hline \hline
    $\{X^l,Y^l\}$  & \xmark & \xmark &\xmark &\xmark & 56.8 & 30.2 & 68.0 & 43.3 & 82.6 & 61.2 \\ \hline
    $\{X^l,Y^l,X^u\}$  & \cmark & \xmark &\xmark & \xmark & 62.8 & 38.4 & 74.0 & 50.6 & 83.4 & 63.7  \\
     $\{X^l,Y^l,X^u\}$ & \cmark & \textit{AP} &\xmark& \textit{Hard} &  78.4 & 55.4 & 83.5 & 63.8  & 88.5 & 71.9\\
    $\{X^l,Y^l,X^u\}$  &\cmark & \textit{AP} &\xmark& \textit{Soft} & 79.0 & 57.0 & 85.5 & 65.9& 88.6 & 72.5    \\
    $\{X^l,Y^l,X^u\}$  &\cmark & \textit{AP} &\cmark& \textit{Soft} & 81.5 & 61.0  & 85.9 & 69.7 & 89.4 & 73.7 \\
    $\{X^l,Y^l,X^u\}$\textbf{(Ours)} & \cmark & \textit{\textbf{SG-AP}} &\cmark& \textit{\textbf{Soft}} & \textbf{87.9} &\textbf{71.6}  &  \textbf{89.8}& \textbf{74.9} &\textbf{91.1} & \textbf{76.4} \\ \hline
    All training data & \multicolumn{4}{c|}{Fully-supervised training} & \multicolumn{6}{c}{ R-1 / mAP : 91.3 / 79.1}\\ \hline
    \end{tabular}
    \vspace{-2mm}
\end{table}

\vspace{-2mm}
\paragraph{\textbf{Model Design of SGC-DPL.~}}
We first assess our initialization strategy in Sec.~\ref{sub:initial} using $\{X^l,Y^l,X^u\}$. As listed in the first two rows in Table~\ref{tab:ablation}, the re-ID model with our initialization strategy outperformed the naive model trained on $\{X^l,Y^l\}$ only, especially on the M-1/12. With initialization confirmed, we next consider assigning hard/soft pseudo-labels to all unlabeled data simply based on standard AP, without semantics guidance from the labeled set. The results are shown in the third and fourth rows in Table~\ref{tab:ablation}, indicating that our soft pseudo-labels can alleviate the errors in AP. From the fifth row of this table, we see that applying our progressive learning strategy for selecting reliable data to augment the labeled training set would help, while replacing the standard AP by our SG-AP would achieve the best results (i.e., our proposed SGC-DPL). Take M-1/12 for example, where only 60 identities are labeled, when comparing to the baseline approach of using soft pseudo-labels by standard AP only (i.e., the fourth row in Table~\ref{tab:ablation}), the performance was increased by a large margin from $\mathbf{57.0}$ to $\mathbf{71.6}$ in mAP, which confirms our ability in jointly exploiting labeled and unlabeled data for improved re-ID learning. Finally, we see that with one third of labels observed (i.e., M-1/3), our model was able to produce comparable performances as the fully-supervised model with the same backbone and same training methods produced by ourselves. (i.e., the last row in Table~\ref{tab:ablation}).

\vspace{-1mm}

\vspace{-2mm}
\paragraph{\textbf{Visualization of SG-AP.}~}
As shown in Fig.~\ref{fig:SG-AP}, we visualize the clustering results across internal searching process in our SG-AP. Data points with the same color represent the same cluster after our SG-AP, while the labeled and unlabeled data are bounded by circles and rectangles, respectively. And, each circle/rectangle indicates a ground truth ID label (e.g., we have $c^l_1$ to $c^l_3$ and $c^u_1$ to $c^u_5$ to denote the ID labels for labeled and unlabeled data, respectively in Fig.~\ref{fig:SG-AP}). 
From the left hand side of this figure, we see that our SG-AP initially divided instances in both labeled and unlabeled data of the same ground truth ID into multiple clusters, which is not desirable. %different clusters, while instances of the same ID in the labeled set were divided into multiple clusters. 
With our SG-AP progresses for searching suitable $p$ described in Sec.~\ref{subsub:semanticAP}, the number of clusters on labeled set would match $C^l$ as shown in the right part of Fig.~\ref{fig:SG-AP}, which also guide the unlabeled ones for improved clustering results (e.g., $c^u_1$ to $c^u_5$ in the right most part in Fig.~\ref{fig:SG-AP}).

\vspace{-3mm}
\subsection{Extension}
\vspace{-2mm}
Although our SGC-DPL mainly tackled the semi-supervised setting practically in the task of person re-ID, it can be generally applied and extended to other tasks with the same setting. Therefore, we extended our SGC-DPL to the tasks of vehicle re-ID on the VeRi-776~\cite{VeRi-776} and image retrieval on the CUB-200~\cite{cub} datasets to verify the generalization ability. Different from other semi-supervised settings, the identities are also disjoint between labeled and unlabeled set in the training data and the ratio of the labeled data is also set as 1/3, 1/6 or 1/12. Compared to person re-ID, vehicle re-ID is a more challenge task owing to the large variation between the same vehicles captured from different views (i.e. rear and front) and the similar appearance between vehicles with the same car model, color and views. The image retrieval on CUB-200 is quite challenging, too. There are only 100 classes in the original training set and we would only have 17 labeled classes if the 1/6 setting is applied. Table~\ref{tab:car} and~\ref{tab:cub} show the promising results that with the guidance of labeled set, our SGC-DPL can compete against the fully-supervised (fully-sup) state-of-the-arts approaches. Details are described in our supplementary materials.
\begin{table}[t!]
    \begin{minipage}{.5\linewidth}
      \centering
      \caption{Comparisons with the state-of-the-arts on VeRi-776~\cite{VeRi-776} (\%).}
        \begin{tabular}{c|c|cc}
        \hline
        Method  & Supervision& R1 & mAP \\
        \hline \hline
        \multirow{4}{*}{\textbf{Ours}} & 1/12 labeled &76.2 &38.0\\
                                       & 1/6 labeled &78.7 &43.6\\
                                       & 1/3 labeled &84.3 &56.7\\
                                       & fully-sup    &90.1 &64.7\\
        \hline \hline
        RAM~\cite{RAM} & \multirow{2}{*}{fully-sup} & 88.6 & 61.5\\
        GRF-GGL~\cite{GRF-GGL}& & 89.4& 61.7 \\
        \hline
     \end{tabular}
     \label{tab:car}
    \end{minipage}
    \begin{minipage}{.5\linewidth}
     \centering
      \caption{Comparisons with the state-of-the-arts on CUB-200~\cite{cub}.}
        \begin{tabular}{c|c|cc}
        \hline
        Method  & Supervision& R1 & NMI~\cite{nmi} \\
        \hline \hline
        \multirow{4}{*}{\textbf{Ours}} & 1/12 labeled & 47.2 & 56.5\\
                                       & 1/6 labeled & 48.1 & 58.7\\
                                       & 1/3 labeled & 48.8 & 59.3\\
                                       & fully-sup    & 49.7 & 59.3\\
        \hline \hline
        Proxy~\cite{proxy} & \multirow{2}{*}{fully-sup} & 49.2 & 59.5\\
        Smart+~\cite{smart}& & 49.8 & 59.9 \\
        \hline
     \end{tabular}
       \label{tab:cub}
    \end{minipage} 
    \vspace{-4mm}
\end{table}
\vspace{-5mm}
\section{Conclusion}
\vspace{-3mm}
In this paper, we presented a novel Semantics-Guided Clustering with Deep Progressive Learning (SGC-DPL) framework for semi-supervised person re-ID. Our core novelty lies in the proposed clustering algorithm, semantics-guided affinity propagation (SG-AP). Without the prior knowledge of the cluster numbers, we are able to cluster unlabeled data with the semantics-preserving guarantees, under the guidance of labeled data. Together with the progressive learning strategy, our model is able to select unlabeled data and assign soft pseudo-ID labels, which allows one to augment the labeled training dataset and thus results in improved re-ID performances. Qualitative and quantitative results confirm the design of our SGC-DPL framework, which performed favorably against recent semi-supervised methods while achieving comparable performances as fully-supervised ones did.
\section{Supplementary Materials}
\begin{table}[t!]
    \centering
    \mycaption{Preliminary experiments with Affinity Propagation and DBSCAN}{This table shows the clustering results with different clustering algorithms and the re-ID performance after training for one iteration.}
    \label{tab:cluster}
    \begin{tabular}{c|c|c|c|cc|c|c|cc}
    \hline
    \multirow{2}{*}{Methods} & \multirow{2}{*}{Param. setting} & \multicolumn{4}{c|}{M-1/3} & \multicolumn{4}{c}{M-1/6} \\
    \cline{3-10}
    && \#cluster & \#ID & R1 & mAP& \#cluster &\#ID & R1 & mAP\\
    \hline \hline
    AP & default & 565 & \multirow{3}{*}{501} & 87.9 & 70.8 & 589 &\multirow{3}{*}{626}& 82.7 & 62.8  \\
    \cline{1-3}\cline{5-7}\cline{9-10}
    DBSCAN & default & 1 &  & -- & -- & 1& & -- & -- \\
    \cline{1-3}\cline{5-7}\cline{9-10}
    DBSCAN & SSG~\cite{SSG} & 381 &  & 88.1 & 70.5 & 266 & & 81.0 & 60.7  \\
    \hline
    \end{tabular}
\end{table}

\subsection{Analysis of the Performance for Different Clustering Algorithms}
Initially, we conduct experiments to evaluate the effectiveness between two widely used clustering solutions that are both no need for deciding the number of clusters in advance, Affinity Propagation (AP)~\cite{AP} and DBSCAN~\cite{DBSCAN}. We did not conduct the experiments of K-means clustering with different K to validate the robustness as in~\cite{icipxin} because we think that even the possible range of the number of identities is also unknown. 
For AP, we all adopt the default hyperparameters proposed in~\cite{AP}. For the hyperparameters in DBSCAN, we adopt two settings, one is with the default values and the other one is proposed in SSG~\cite{SSG}. Experiments are conducted on the unlabeled set of M-1/3 and M-1/6, whose feature extractors are only initialized on each $\{X^l,Y^l\}$, respectively. Table~\ref{tab:cluster} shows the results. We demonstrate the number of cluster in the first iteration and its ground truth number of identities. In addition, we also show the re-ID performance after the first training iteration ($t=1$) with hard pseudo-labels and without progressive learning on $\{X^l,Y^l,X^u,Y^p\}$. It can be seen that with the default setting in DBSCAN, we obtain an undesirable clustering results. With the meticulous design in DBSCAN that follows SSG, the performance can just compete against the AP with default values. Thus, in our SGC-DPL, we choose to adopt AP for clustering the unlabeled data.

\subsection{Ablation Studies on DukeMTMC-reID~\cite{DukeReID}}
Due to page limit of the main paper, we also provide the ablation studies on the other large-scale dataset, DukeMTMC-reID~\cite{DukeReID} (D), with three semi-supervised settings depend on the ratio of labeled identities (1/3, 1/6 and 1/12) and thus denoted as D-1/3, D-1/6, and D-1/12 in Table~\ref{tab:duke_ablation}. By the way, we did not apply our method on MSMT17 dataset~\cite{msmt} because it is no longer available now. Same as the comparison on Market-1501~\cite{Market1501} in the main paper, it also shows the effectiveness of our proposed SGC-DPL framework. When comparing to the baseline method using soft pseudo-labels by standard AP only (i.e., the fourth row in Table.~\ref{tab:duke_ablation}), the performance on D-1/12 was increased by a large margin, too. Furthermore, the performance of our SGC-DPL on D-1/3 can also approach that with the fully-supervised method.

\begin{table}[t!]
    \centering
    \mycaption{Ablation studies of the proposed method on DukeMTMC-reID in terms of R-1 and mAP ($\%$)}{Note that the settings are the same as those in the main paper. All methods in this table share the same backbone model.}
    \label{tab:duke_ablation}
    \ra{1}
    \begin{tabular}{l|cccc|cc|cc|cc}
    \hline
    \multirow{2}{*}{Experimental setting} & \multicolumn{4}{c|}{Components} & \multicolumn{2}{c|}{D-1/12}&\multicolumn{2}{c|}{D-1/6}&\multicolumn{2}{c}{D-1/3} \\
    \cline{2-11}
     & Init. & Clus. & P.L. & Pseu.-labels& R-1 & mAP & R-1 & mAP & R-1 & mAP \\
    \hline \hline
    \{$X^l,Y^l\}$ & \xmark & \xmark &\xmark &\xmark & 46.2 & 27.0 & 61.2 & 40.6 & 71.5 & 53.1  \\
    $\{X^l,Y^l,X^u\}$ & \cmark & \xmark &\xmark & \xmark & 50.0 & 29.1 & 65.0 & 43.5 & 73.7& 54.9\\
    \hline
    $\{X^l,Y^l,X^u\}$ & \cmark & \textit{AP} &\xmark& \textit{Hard}& 63.3 & 44.0 & 73.2 & 55.4& 79.0 & 62.0    \\
    $\{X^l,Y^l,X^u\}$ &\cmark & \textit{AP} &\xmark& \textit{Soft} & 65.9 & 47.1 & 73.8 & 55.8 & 79.4 & 62.4 \\
    $\{X^l,Y^l,X^u\}$ &\cmark & \textit{AP} &\cmark& \textit{Soft} & 68.9 & 50.9 & 74.7 & 57.3 &78.7 & 63.3  \\
    $\{X^l,Y^l,X^u\}$\textbf{(Ours)} & \cmark & \textit{\textbf{SG-AP}} &\cmark& \textit{\textbf{Soft}} &\textbf{74.1} &\textbf{56.4} & \textbf{77.6} & \textbf{61.0}  & \textbf{82.2} &\textbf{66.5} \\ \hline
    All training data & \multicolumn{4}{c|}{Fully-supervised training} & \multicolumn{6}{c}{R-1 / mAP : 85.5 / 71.3 }\\ \hline
    \end{tabular}
\end{table}

\begin{figure}[t!]
	\centering
    \includegraphics[width=\textwidth]{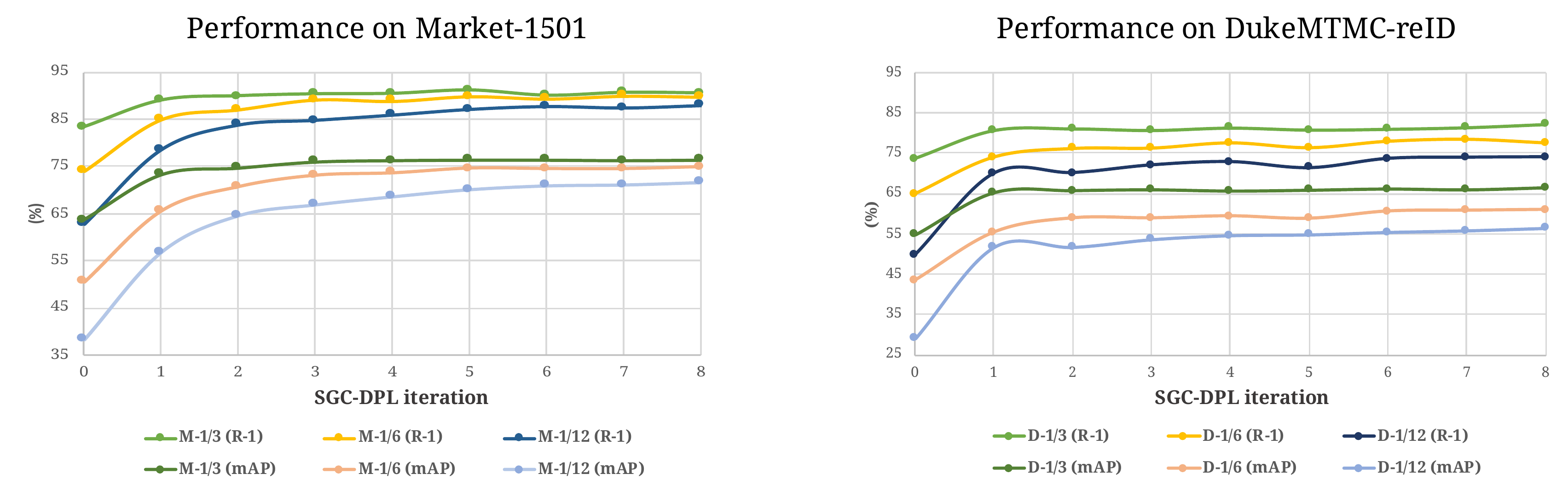}
    \mycaption{Performance on two datasets along the SGC-DPL iteraions}{We see that the performances generally converged after the $5^{th}$ iteration. Thus, we had $t=8$ in our work which would be a reasonable choice.}
    \label{fig:iteration}
\end{figure}

\subsection{Analysis of total \#iterations in SGC-DPL}
We analyze the hyper-parameter $t$, which is the total number of iterations in our SGC-DPL framework. Fig.~\ref{fig:iteration} shows the performance along the SGC-DPL iterations in terms of rank-1 and mAP on Market-1501 and DukeMTMC-re-ID datasets, respectively. Each dataset consists of three semi-supervised settings considered. The $0^{th}$ iteration represents the model performance after our initialization method. From these figures, we observe that the performance converged after the $5^{th}$ iteration in both datasets. Thus, we set $t=8$ which would be a reasonable choice for the proposed SGC-DPL framework.

\begin{figure}[t]
	\centering
    \includegraphics[width=0.95\textwidth]{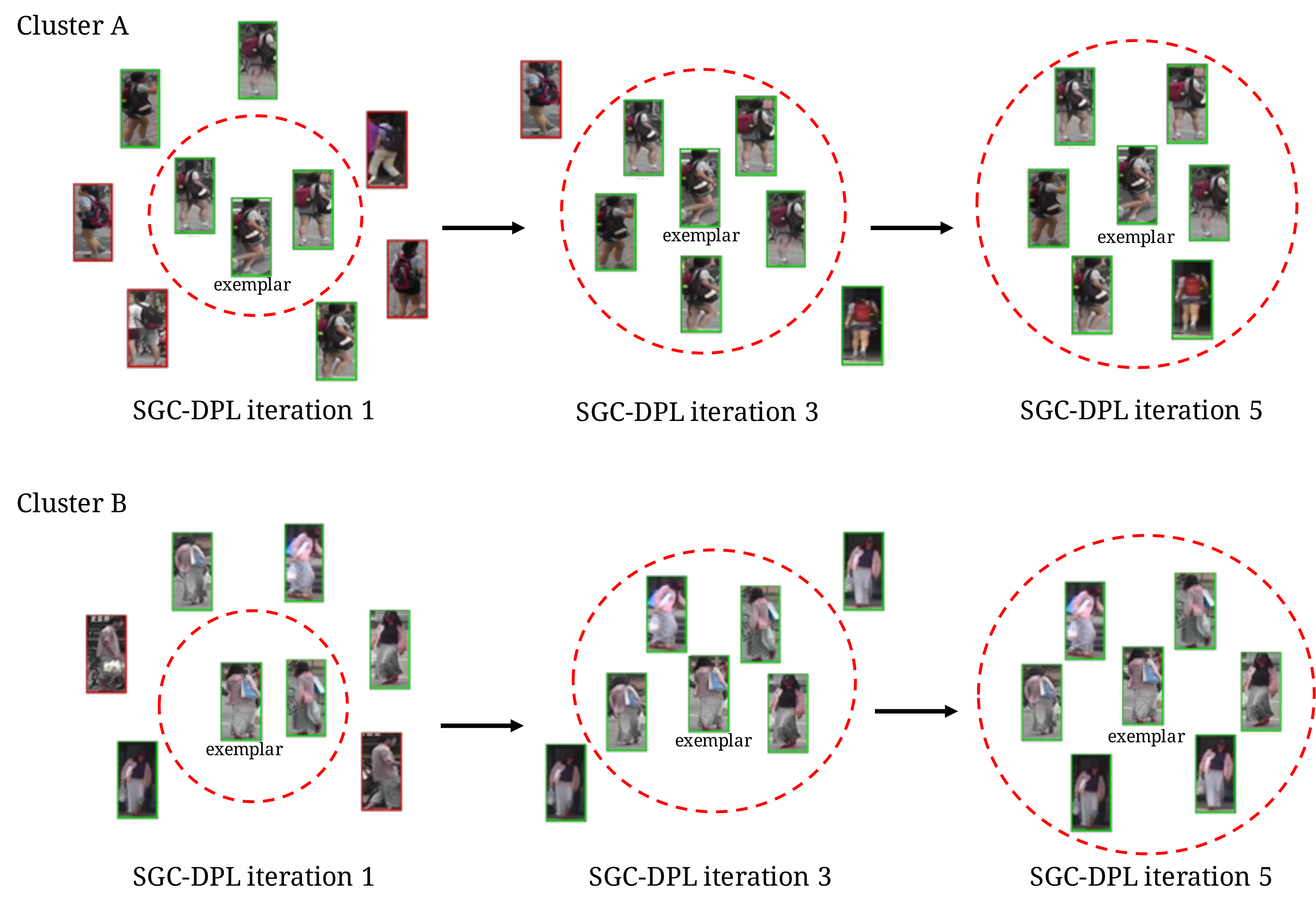}
    \mycaption{Visualization of our progressive learning strategy on M-1/6}{We illustrate example results of selected two clusters by SGC-DPL. The images in green bounding boxes represent those with the same ID (as that of the cluster exemplar), while images in red bounding boxed are not. The red dotted circle denotes the reliable data subset selected. We see that the ID labels were noisy in the beginning of clustering. Reliable data selected over iterations would update both pseudo-label prediction and clustering, which effectively augment labeled data from unlabeled data for improved learning.}
    \label{fig:PL}
    \vspace{-3mm}
\end{figure}

\subsection{Visualization of our Progressive Learning Strategy}
For each iteration in our progressive learning strategy, we will create a reliable subset for each cluster with the threshold $\tau$ which is automatically generated based on the labeled set. The $\tau$ will be increased progressively in each iteration to enlarge the subset till all the samples are in the subset. Fig.~\ref{fig:PL} shows two visualized cluster examples in our SG-DPL iterations on M-1/6 dataset. Each row represents the cluster members of the same exemplar along the iterations. The images with green border are with the same ground truth identities to the exemplar, and those with red are not. The red circle represents the reliable subset. We can observe that for the first iteration, the cluster results contain some errors which includes the incorrect identities. However, with our threshold for the reliable subset, we only assign pseudo-labels to the correct samples. As the network be optimized on the correct data, the cluster results will be more accurate. Furthermore, as the threshold be enlarged, more correct data will be assigned pseudo-labels for learning re-ID model.

\subsection{Implementation details of the extension experiments}

For vehicle re-identification, the widely used VeRi-776 dataset~\cite{VeRi-776} contains 776 different vehicles captured, which is split into 576 vehicles with 37,778 images for training and 200 vehicles with 11,579 images for testing. The training details all follow those in our main paper for person re-ID, which contains the same CNN backbone and the same three training tricks proposed in~\cite{bagoftricks}.

For image retrieval, we adopt CUB-200 dataset~\cite{cub}. This dataset is a fine-grained bird dataset containing 11,788 images of 200 bird species. Following existing methods~\cite{proxy,smart}, we use the first 100 categories with 5,864 images for training, and the remaining 100 categories with 5,924 images for testing. The ratio for the labeled data in our semi-supervised setting is also applied on the 100 training classes. For learning on labeled or pseudo-labeled data, we follow the triplet training network proposed in~\cite{smart}. The reason for choosing~\cite{smart} but not other state-of-the-arts is that adopting this purely triplet training can easily demonstrate the performance improvement with or without our SGC-DPL. In Table~3 \& 4 of our main paper, the performances of the ``fully-sup'' setting produced by ourselves are the upper-bound of our SGC-DPL method on two datasets, which means we directly train the supervised network with all training data. 
%===========================================================
\bibliographystyle{splncs04}
\bibliography{egbib}

%this would normally be the end of your paper, but you may also have an appendix
%within the given limit of number of pages
\end{document}